\documentclass[letterpaper, 10 pt, conference]{ieeeconf}  

\IEEEoverridecommandlockouts                              

\usepackage{mathptmx} 
\usepackage{times} 
\usepackage{amsmath} 
\usepackage{amssymb}  
\usepackage{subcaption}
\let\labelindent\relax
\usepackage{enumitem}
\usepackage{diagbox}
\usepackage{bbding}
\usepackage{booktabs}
\usepackage{pifont}
\usepackage{wrapfig}
\usepackage{lipsum}
\usepackage{multirow}
\usepackage{url}
\usepackage{xcolor}
\usepackage{graphicx}
\usepackage{algorithm}
\usepackage{algpseudocode}

\title{\LARGE \bf
Learning Generalizable Tool-use Skills through Trajectory Generation
}
\newcommand{\fig}[1]{Fig.~\ref{#1}}  
\newcommand{\tbl}[1]{Table~\ref{#1}} 
\newcommand{\algo}[1]{Algorithm~\ref{#1}} 
\newcommand{\sect}[1]{Sec.~\ref{#1}}

\author{ Carl Qi$^{*1}$
  Yilin Wu$^{*2}$ 
  Lifan Yu$^{2}$ 
  Haoyue Liu$^{2}$
  Bowen Jiang$^{2}$
  Xingyu Lin$^{**3}$
  David Held$^{**2}$ \\
\thanks{$^{1}$University of Texas at Austin, United States}%
\thanks{$^{2}$Carnegie Mellon University, United States}%
\thanks{$^{3}$University of California, Berkeley, United States}%
\thanks{$^{*}$ equal contribution}%
\thanks{$^{**}$ equal advising}
}

\begin{document}
\maketitle

\begin{abstract}
Autonomous systems that efficiently utilize tools can assist humans in completing many common tasks such as cooking and cleaning. However, current systems fall short of matching human-level of intelligence in terms of adapting to novel tools. Prior works based on affordance often make strong assumptions about the environments and cannot scale to more complex, contact-rich tasks. In this work, we tackle this challenge and explore how agents can learn to use previously unseen tools to manipulate deformable objects. We propose to learn a generative model of the tool-use trajectories as a sequence of tool point clouds, which generalizes to different tool shapes. Given any novel tool, we first generate a tool-use trajectory and then optimize the sequence of tool poses to align with the generated trajectory. We train a \textit{single model} on four different challenging deformable object manipulation tasks, using demonstration data from only one tool per task. The model generalizes to various novel tools, significantly outperforming baselines. We further test our trained policy in the real world with unseen tools, where it achieves the performance comparable to human.
    Additional materials can be found on our project website.\footnote{\url{https://sites.google.com/view/toolgen}}
\end{abstract}


\section{INTRODUCTION}
Building autonomous systems that leverage tools can greatly enhance efficiency and assist humans in completing many common tasks in everyday life~\cite{xie2019improvisation, fang2020learning, qin2020keto, turpin2021gift, lin2021diffskill, qi2022learning, lin2022planning}. As humans, we possess an innate ability to adapt quickly to use novel tools.
However, replicating such adaptability in autonomous systems presents a significant challenge. 
To solve this task of novel tool manipulation, prior work has explored different representations for tools. A good tool representation should contain a rich visual understanding of the object and be useful for downstream physical interactions. Prior work~\cite{fang2020learning} uses data-driven approaches to learn the latent representations for tools but such representation cannot generalize because of the lack of compositionality and interpretability. Another line of the work studies keypoints as a representation for tool which works only for rigid object manipulation including hammering, pushing and reaching.

In this work, we explore how agents can learn to use novel tools to manipulate deformable objects. Beyond the challenges of representing novel tools, manipulating deformable objects with tools adds considerable difficulties. For one, manipulating deformable objects often results in rich, continuous contact between the tool and the object; the contacts between a roller tool and dough, for example, are continuous and cannot be easily discretized, which makes specifying discrete affordance labels to describe such interactions difficult. 
Further, defining rewards or keypoints (as is sometimes used for tool and environment representations~\cite{qin2020keto, turpin2021gift})
for deformable objects is also challenging. 
Therefore, operating novel tools to solve diverse tasks calls for an approach that makes few assumptions about the task and the environment.
\begin{figure}[t]
    \centering
    \includegraphics[width=\linewidth]{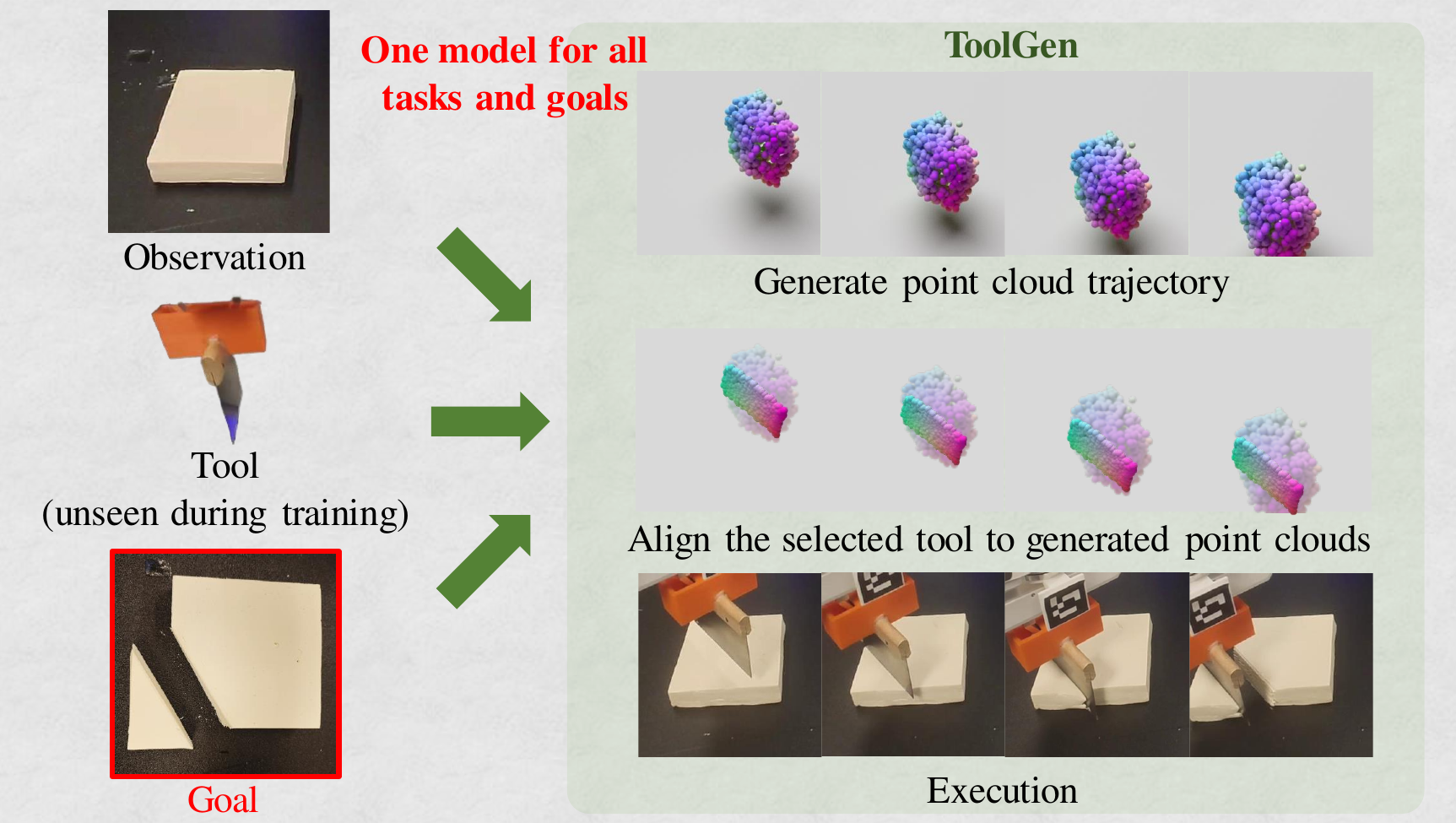}
    \caption{Our method ToolGen can solve deformable object manipulation with diverse tasks and goals. It does so by first generating a point cloud trajectory of the desired tool and then aligning the actual tool to the generated point clouds for execution. We train a single model for four different challenging deformable object manipulation tasks. Our model is trained with demonstration data from just a single tool for each task and is able to generalize to various unseen tools.
    }
    \label{fig:pull}
    \vspace{-0.7cm}
\end{figure}
 Our goal is to train a 
 policy to solve various manipulation tasks with multiple tools, including tools that were not seen during training. We propose a novel approach, ToolGen, which learns tool-use skills via trajectory generation and sequential pose optimization. 
 Given the scene, the goal, and a tool, ToolGen first
 generates a point cloud of a tool in the desired initial pose, and it subsequently predicts how this generated tool would move to perform the task. Finally, we sequentially align the actual tool to the generated tool to extract the actions for the agent to execute. Fig~\ref{fig:pull} offers an overview of our task setting and ToolGen's outputs. We evaluate ToolGen against several baselines in deformable object manipulation with diverse tasks, goals, and tools. Impressively, with just a single model trained across all tasks and tools, ToolGen significantly outperforms the baselines and generalizes to many novel tools. Further, ToolGen achieves this despite being trained on demonstrations from just one tool for each task. 
 
 To summarize our contribution, we propose ToolGen, which represents tool use via trajectory generation. We have shown that generating a point cloud trajectory of the tool can effectively capture the essence of tool use, i.e. how the tool should be placed in relation to the dough and how it should move over time, which allows us to generalize to a variety of unseen tools and goals. Furthermore, we transfer the policy to the real world without any finetuning to demonstrate our method's effectiveness on three real world manipulation tasks with unseen novel tools and different goals. 


\begin{figure*}[ht]
    \centering
    \includegraphics[width=\linewidth]{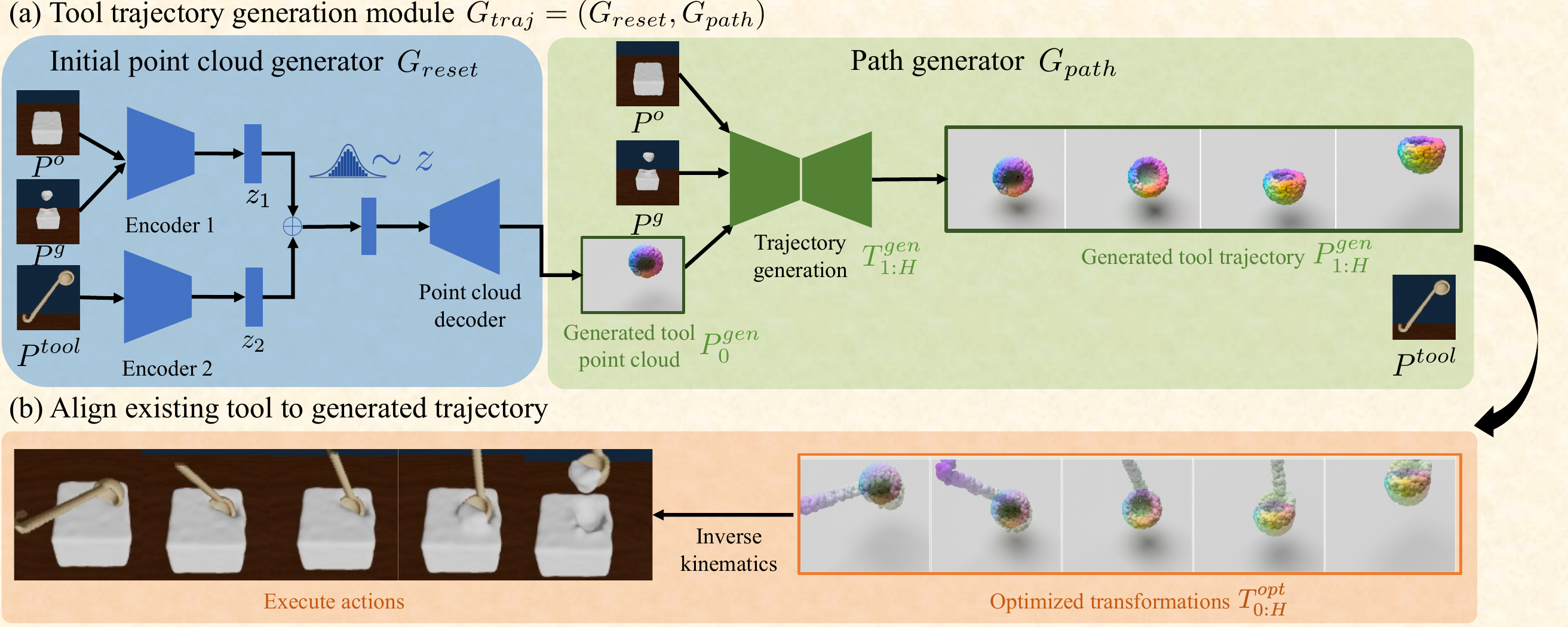}
    \caption{Overview of our method: (a) Given an initial observation of the scene $P^o$, the goal $P^g$, and a tool $P^{tool}$, we first leverage the trajectory generation module $G_{traj}$ to generate an ideal tool trajectory accomplishing the task $P^{gen}_{0:H}$. It encompasses two submodules: Initial point cloud generator $G_{reset}$ generating reset pose $P^{gen}_0$ of reconstructed tool and Path generator $G_{path}$ generating $P_{1:H}^{gen}$ (b) We then align the existing tool with the reconstructed tool via sequential pose optimization to extract the pose of the existing tool $T^{opt}_{0:H}$, and we subsequently use inverse kinematics to obtain the actions for the agent to execute.}
    \label{fig:overview}
\end{figure*}
\section{Related Work}
\textbf{Learning Generalizable Tool-use Skills:}  Prior works have explored training robots to perform manipulation tasks with tools. To enable generalization, some approaches predict intermediate ``affordances'' and then generate actions based on these affordances~\cite{fang2020learning, zhu2015understanding, manuelli2019kpam, gao2021kpam}. For example, affordances like grasping points or functional points and be represented as key points~\cite{fang2020learning, qin2020keto,turpin2021gift, manuelli2019kpam, gao2021kpam}. Similarly, concepts like contacts and forces~\cite{zhang2022understanding,wi2022virdo++} can also be used. However, obtaining labels for these affordances can be difficult, and such affordance labels do not easily extend to deformable object manipulation, since the contacts with deformable objects (e.g. rolling a piece of dough) are complex and cannot be modeled by a few keypoints. Comparing to these methods, our method is capable of learning from unlabeled interaction data, as it implicitly learns affordances from the point clouds of the tool and the dough. This data-driven approach is similar to prior work~\cite{thompson2021shape}, but we do not explicitly specify the structure of the shape embedding space, leaving more flexibility in tool shapes.

Another approach is to discover affordance regions in a self-supervised way by running parameterized motion primitives~\cite{fang2020learning} or affordance-conditioned policies~\cite{qin2020keto,turpin2021gift} in simulation. 
In the image space, prior works have explored training an action-conditioned video prediction model~\cite{xie2019improvisation} for planning actions for different tools. However, the video prediction model lacks 3D structure and has difficulty representing fine-grained action trajectories. Another research direction for generalizable tool use is to utilize the pretrained Large Language Models (LLMs) for long horizon reasoning. Prior work~\cite{xu2023creative} designs four consecutive modules to prompt LLMs to directly generate code for robotic tool use. However, they make an assumption that we have state information of the tools and objects and directly include them into the prompt for LLMs. For deformable objects, state estimation is very challenging so this approach doesn't generalize to deformable object manipulation with tool use.  

\textbf{Deformable Object Manipulation with Tools:} 
Prior works with deformable objects often consider using a fixed set of tools. For example, Some approaches~\cite{lin2021diffskill,lin2022planning} aim to solve the task of dough manipulation with a differentiable simulator but their tool sets are fixed with rolling pin and spatula. Other work~\cite{heiden2021disect,xu2023roboninja} use a fixed tool set of knives for cutting. These works do not consider generalization to novel tools, which is the focus of this work. 

 






\section{Problem statement and assumptions}
\label{sec:problem_statement}
Consider a set of point clouds $(P^o, P^g, P^{tool})$, where $P^o$ represents the initial observation of the scene, $P^g$ stands for the goal, and $P^{tool}$ for a tool to use for execution. Our task is to predict an actions sequence of horizon $H$, where the tool transforms the initial pose into a predicted target pose. The actions is represented by a transformations sequence $T_{0:H}$. Here, all the point cloud positions as well as the objects' orientations are relative to a reference frame located at the dough center. This design allows us to perform manipulation that is agnostic to the location of the dough on the table.

In the training stage, we use demonstrations of tools from a training set $\{P^{traintool_i}\}_{i=1:K_{train}}$, where $K_{train}$ is the number of training tools we have. The demonstration data fed into the model are of the form: $(P^o, P^g, P^{traintool_i}, T_{0:H})$. The initial transformation $T_0$ in the sequence brings the tool to a ``reset pose''. The remaining terms $T_{1:H}$ are the relative transformations from the previous timestep, which we call ``delta poses.'' For each task, we manually specify distributions of the initial and goal configurations. We then run trajectory optimization using a differentiable simulator to generate these demonstrations following prior works~\cite{huang2021plasticinelab}. Human demonstrations could serve as an alternative source of the training data described above.

\section{Method}

We propose the following approach to obtain an trajectory executable for robots with any given tools: 
\begin{itemize}
    \item We first generate a point cloud of a reconstructed tool $P^{gen}$ at a starting pose based on the given tool $P^{tool}$ (\sect{sec:gen_traj_gen}).
    \item Next we generate a sequence of tool actions of how this generated tool would achieve the task (\sect{sec:gen_traj_gen}) based on policy learned with Behavior Cloning. 
    \item We then align the actual tool to each of the point clouds in the generated trajectory (\sect{sec:real_traj_gen}). 
\end{itemize}
Below, we describe this approach in detail, and experiments in \sect{sec:results} demonstrate the remarkable improvements of this approach compared to other approaches.

\subsection{Representing tool-use through point cloud trajectory generation} \label{sec:gen_traj_gen}

In this section, we describe our approach for trajectory generation. A straightforward method for trajectory generation would be to directly predict the motion of the tool.
However, directly regressing into the tool's pose, particularly the orientation, is proved to be challenging, as indicated by prior studies~\cite{peretroukhin2020smooth,zhou2019continuity,chen2022projective}. To alleviate this challenge, we employ a generative module $G_{traj}$ to produce a point cloud trajectory $P^{gen}_{0:H}$ to complete the task with reconstructed tool. Our trajectory generation model consists of two parts. In the first part, a initial point cloud generator $G_{reset}$ is utilized to reconstruct a tool point cloud at ``reset pose''. In the second part, a path generator $G_{path}$ is adopted for producing trajectory of $P^{gen}_{0:H}$ based on the reconstructed tool. This generated trajectory will later be used to determine the actions of the actual tool.

$G_{reset}$ is a PointFlow-based~\cite{pointflow} encode-decoder generation model. It conditions on the point cloud of the existing tool $P^{tool}$, the initial scene observation $P^o$, and the goal $P^g$, to reconstruct the tool at ``reset pose'', $P^{gen}_0$. The architecture of our PointFlow-based~\cite{pointflow} generator $G_{reset}$ is shown in~\fig{fig:overview} (a) (top). It encodes the tool points and the concatenation of initial and target dough points to two sets of latent features with separate PointNet++~\cite{pn2} encoders. These latent features are concatenated and inputted through an MLP to produce an estimation of Gaussian distribution. We then take a sample from this estimated distribution as the input of a PointFlow~\cite{pointflow} decoder, which outputs the reconstruction point cloud of the given tool at the reset pose $P^{gen}_0$.

The second part $G_{path}$ works on predicting a sequence of transformations of how this generated tool would move to achieve the task. The architecture of the path generator is shown in~\fig{fig:overview} (a) (top right).
We follow the design in ToolFlowNet~\cite{seita2023toolflownet} to train a policy model through Behavior Cloning, which optimizes a combined loss of point content loss and consistency loss. The $P^{gen}_0$ from $G_{reset}$ is concatenated together with the initial scene observation $P^o$, and the goal state $P^g$ and passed into the model. Transformations of $H-1$ time-steps, $T^{gen}_{1:H}$, are generated for the tool. Details for $G_{path}$ can be found in Appendix~\ref{app:pathgen} on the website.

Together, our generative module $G_{traj} = (G_{reset}, G_{path})$ predicts a trajectory of point clouds $P^{gen}_{0:H}$, which shows the movement of a reconstructed tool accomplishing the manipulation task. Training details are described in Section~\ref{sec:implementation}.

\subsection{Execution via sequential pose optimization} \label{sec:real_traj_gen}

In Section~\ref{sec:gen_traj_gen}, the generated point cloud trajectory of the tool $P^{gen}_{0:H}$ is built upon the reconstructed tool and is not guaranteed to be executable for the actual tool. In this section, we describe the optimization procedure for aligning the actual tool with the generated tool reconstruction in order to extract reasonable actions for actual execution (visualized in \fig{fig:overview} (b) and listed in detail in \algo{alg:opt}). 

The initial transformation at time-step 0 exerts a decisive influence on the overall trajectory. We therefore subdivide the optimized transformations $T^{opt}_{0:H}$ into the reset transformation  $T^{opt}_0$ and delta pose optimization $T^{opt}_{1:H}$.
To align the actual tool $P^{tool}$ to the reconstructed tool in the first timestep $P^{gen}_0$, we consider the following terms: 1) the similarity between the predicted reset pose and actual tool pose, 2) the collision between tool and the initial scene observation. 
The loss function is given by:
\begin{equation}
\begin{aligned}
    \label{eq:reset_metric}
    J_{reset}(T) = Chamfer(T \circ P^{tool}, P^{gen}_0) \\
    - \lambda_{c} \cdot Chamfer(T \circ P^{tool}, P^o),
\end{aligned}
\end{equation}
The first term is the Chamfer distance between the actual tool $P^{tool}$ transformed by $T$ and the reconstructed tool $P^{gen}_0$ at reset pose. The second term is a penalty term computed as the Chamfer distance between the existing tool $P^{tool}$ transformed by $T$ and the observation of the dough $P^o$. $\lambda_c$ is a hyper-parameter balancing the two terms. The aim of the penalty term is to avoid undesirable collisions between the tool in reset pose and the environment, while collisions will be allowed for subsequent time-steps.

For optimization, we use Projected Gradient Descent, detailed in \sect{sec:implementation}, for different initializations of $T$ and learn to start from the one that minimizes the objective described in Eq.~\ref{eq:reset_metric}.

\begin{algorithm}
\caption{Sequential pose optimization}
\begin{algorithmic}[1]
\State \textbf{Input:} The current observation of the dough $P^o$, the existing tool $P^{tool}$, and the point cloud trajectory for the generated tool $P^{gen}_{0:H}$
\State \textbf{// Optimize for the reset transformation}
\State Initialize random transformations $T_0^1, ..., T_0^N$ in $SE(3)$;
\State Optimize $T_0^1, ..., T_0^N$ according to Eq. 1 to obtain costs $J_{reset}^1 ... J_{reset}^N$;
\State Choose the transformation that minimizes the costs, denoted as $T_{0}^{opt}$;
\State \textbf{// Optimize for delta poses}
\State Initialize the delta poses as identities, i.e., $T_{1:H} = I$;
\State Optimize the delta poses according to Eq. 2 and obtain the final transformations $T_{1:H}^{opt}$;
\State \textbf{Output:} Optimized transformations for the existing tool: $T_{0:H}^{opt}$
\end{algorithmic}
\label{alg:opt}
\end{algorithm}

Next we work on the optimization of the delta poses $ T^{opt}_{1:H}$. Similar to that for reset pose, we evaluate the distance between the actual tool $P^{tool}$ and the reconstructed tool at each time-step $P^{gen}_t$, with an additional penalty term to encourage small motions.
The loss function for the delta poses is given by:
\begin{equation}
    \label{eq:delta_metric}
    \begin{aligned}
        J_{\delta}(T_{1:H}) = \sum_{t=1:H} Chamfer( T_{t} \circ X_{t-1}\circ P^{tool}, P^{gen}_t) + \lambda_r \cdot \|T_t \| \\
    \text{where}~X_{t-1} = T_{t-1} \circ T_{t-2} \circ ... T^{opt}_0
    \end{aligned}
\end{equation}
The first term is the Chamfer distance between the reconstructed tool points $P^{tool}$ transformed by $T_{t} \circ X_{t-1}$ and the generated tool points $P^{gen}_t$ at time-step $t$, $\|\cdot \|$ is a regularization function to moderate the magnitude of the translation and rotation defined by the delta poses (see~\sect{sec:implementation} for details). $\lambda_r$ is a hyper-parameter balancing the two terms.

Finally, we apply these objectives in an optimization routine, as outlined in \algo{alg:opt}, to align the reconstructed tool with the generated one and produce the final trajectory $T^{opt}_{0:H}$ for the reconstructed tool. Subsequently, we can utilize inverse kinematics to determine the required actions for our agent to execute the task. In our case, these actions comprise the translation and angular velocities of the tool.

       

\begin{figure}[ht]
    \centering
    \includegraphics[width=\linewidth]{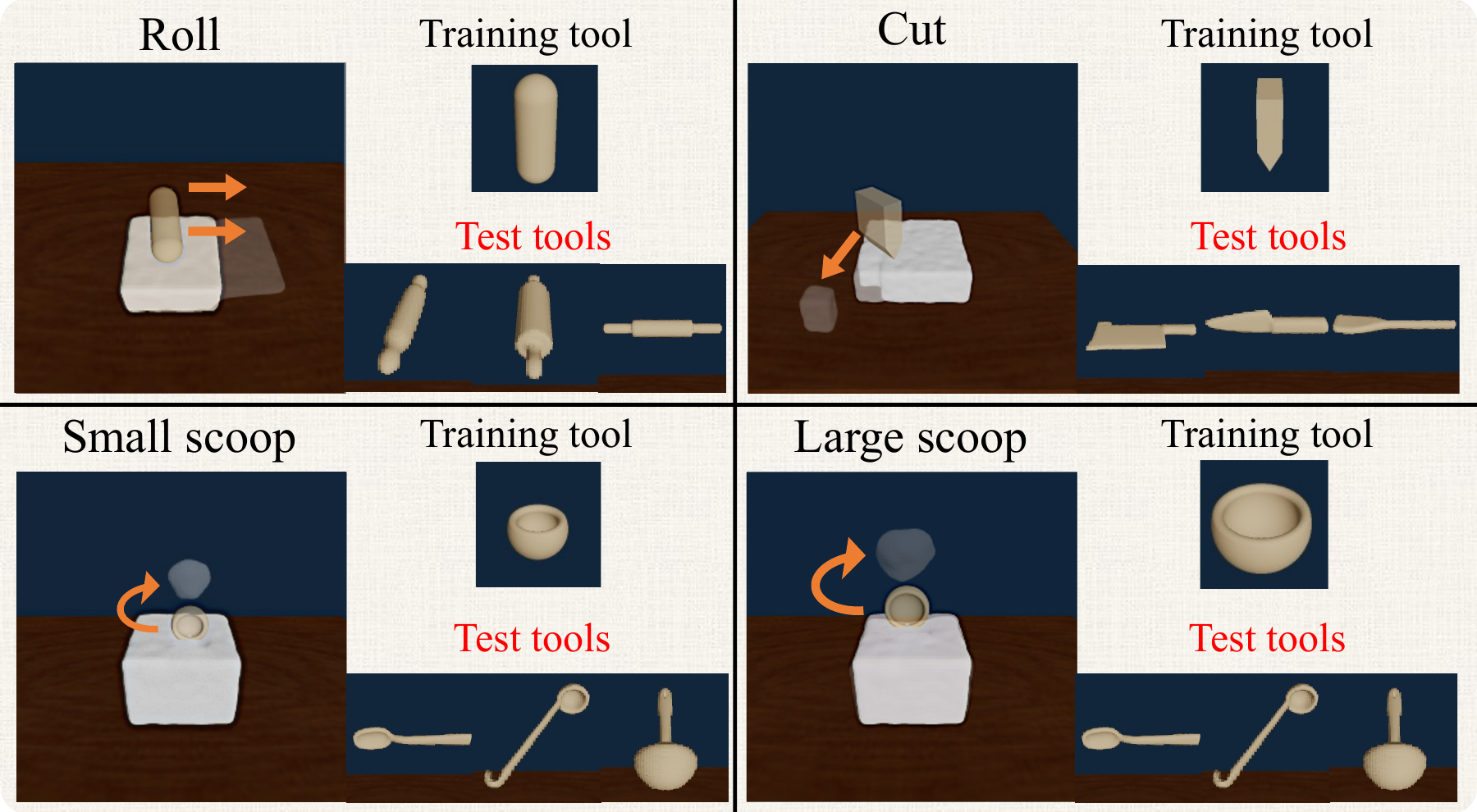}
    \caption{We consider 4 tasks: Roll, Cut, Small scoop, and Large scoop. On the left side of each task, we illustrate how the training tool is used to achieve the goal, overlaying the goal on the initial observation. On the right side, we visualize the initial configurations of the training tool and test tools for each task, highlighting the ability of our method to generalize to novel tools.}
    \label{fig:tasks}
    \vspace{-4mm}
\end{figure}
\subsection{Implementation details}
\label{sec:implementation}
Before imputting the tool, the dough, and the goal point clouds into the PointNet++ networks, we use a one-hot encoding to differentiate points that belong to different objects. Therefore, the input features per point will be $[x, y, z, \text{one-hot}]$.

The two modules of trajectory generation, $G_{reset}$ and $G_{path}$, are trained separately. $G_{reset}$ learns by optimizing the evidence lower bound (ELBO) given the training tools and their reset poses $T_0 \circ P^{traintool_i}$ from the demonstration dataset described in Sec.~\ref{sec:problem_statement}. The trajectories of the training tools $T_{1:H}$ from the demonstration dataset described in Sec.~\ref{sec:problem_statement} are then used as labels for the training $G_{path}$. 

We train a single set of modules ($G_{reset}, G_{path}$) across a compact demonstration dataset comprised of multiple tasks rather than training separate networks for each task. To achieve this, we introduces a scoring module $D_{score}$ to evaluate and select tool for each task. Details of $D_{score}$ module is shown in Appendix~\ref{app:toolsel} on the website. In training data, we collect 200 demonstration trajectories for each task performed with just one training tool. Despite the limited training data, our model is demonstrated to be capable of generalizing to various unseen tools in both simulation and real world. See Appendix~\ref{app:dataset} on the website for more information on our demonstration dataset.

In trajectory optimization, we use the quaternion representation for the orientation of the transformation, and project the values onto a unit ball after each gradient update. Here, we use a step size of $10^{-2}$, and $\lambda_c=0.1$. For optimizing the delta poses, we use the 3-DoF Euler angles representation with a step size of $10^{-3}$, a regularization factor of $\lambda_r=0.1$, and we use the euclidean norm to regularize the translation as well as the rotation. We use a greedy IK solver~\cite{zhang2020modular} to obtain the robot actions from the simulation and the real world, and we find it to work well in our tasks. One could also use IK as the main objective in the sequential pose optimization step to produce better poses for IK to solve.

\begin{figure*}[t]
\centering
\begin{subfigure}[b]{.45\textwidth}
  \centering
  \includegraphics[width=\linewidth]{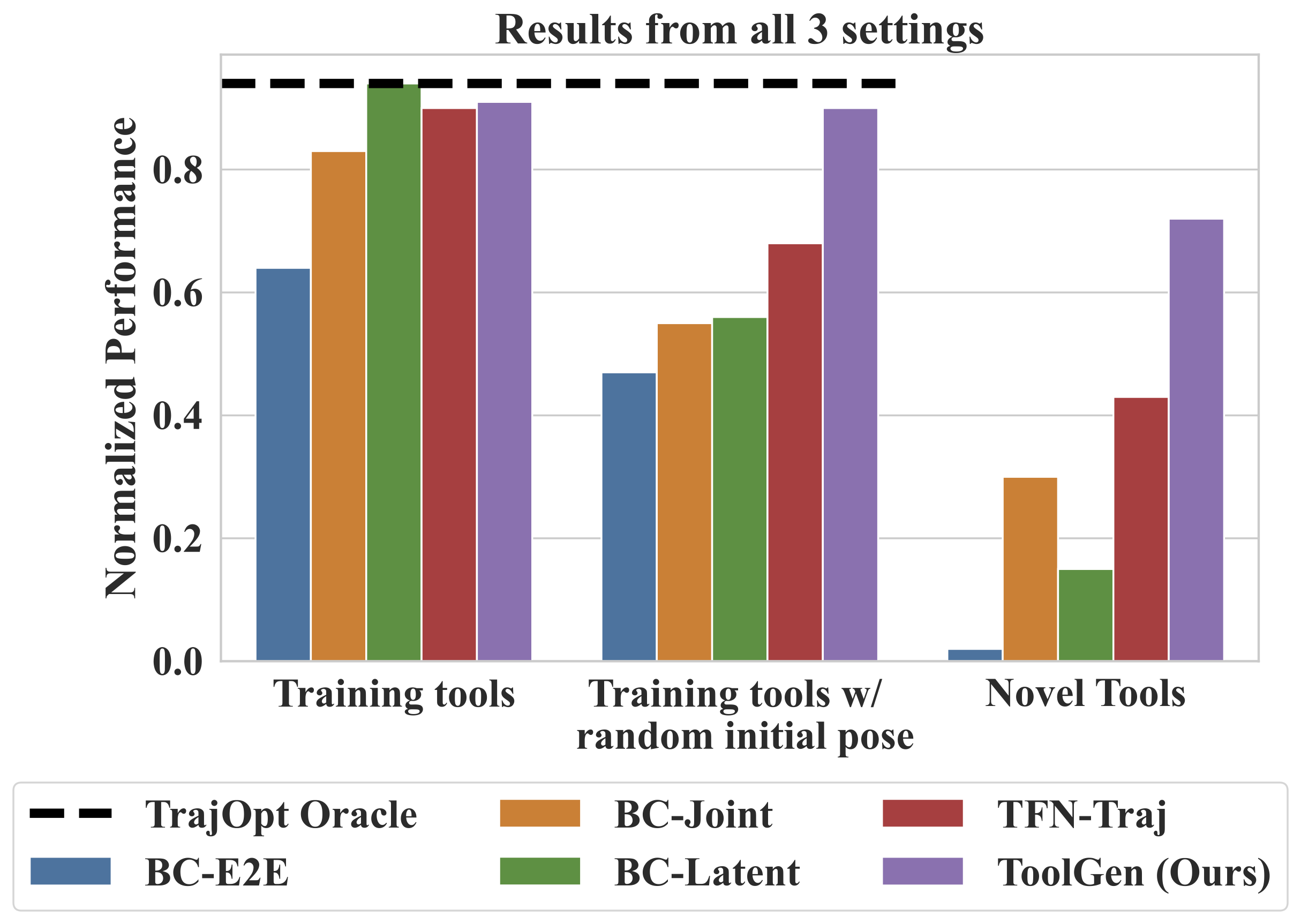}
  \caption{Performance of all the methods across 3 different settings. We evaluate 10 trajectories per task per tool and then aggregate the performance across all the tasks.}
  \label{fig:results}
\end{subfigure}
\hfill
\begin{subfigure}[b]{.5\textwidth}
  \centering
  \includegraphics[width=\linewidth]{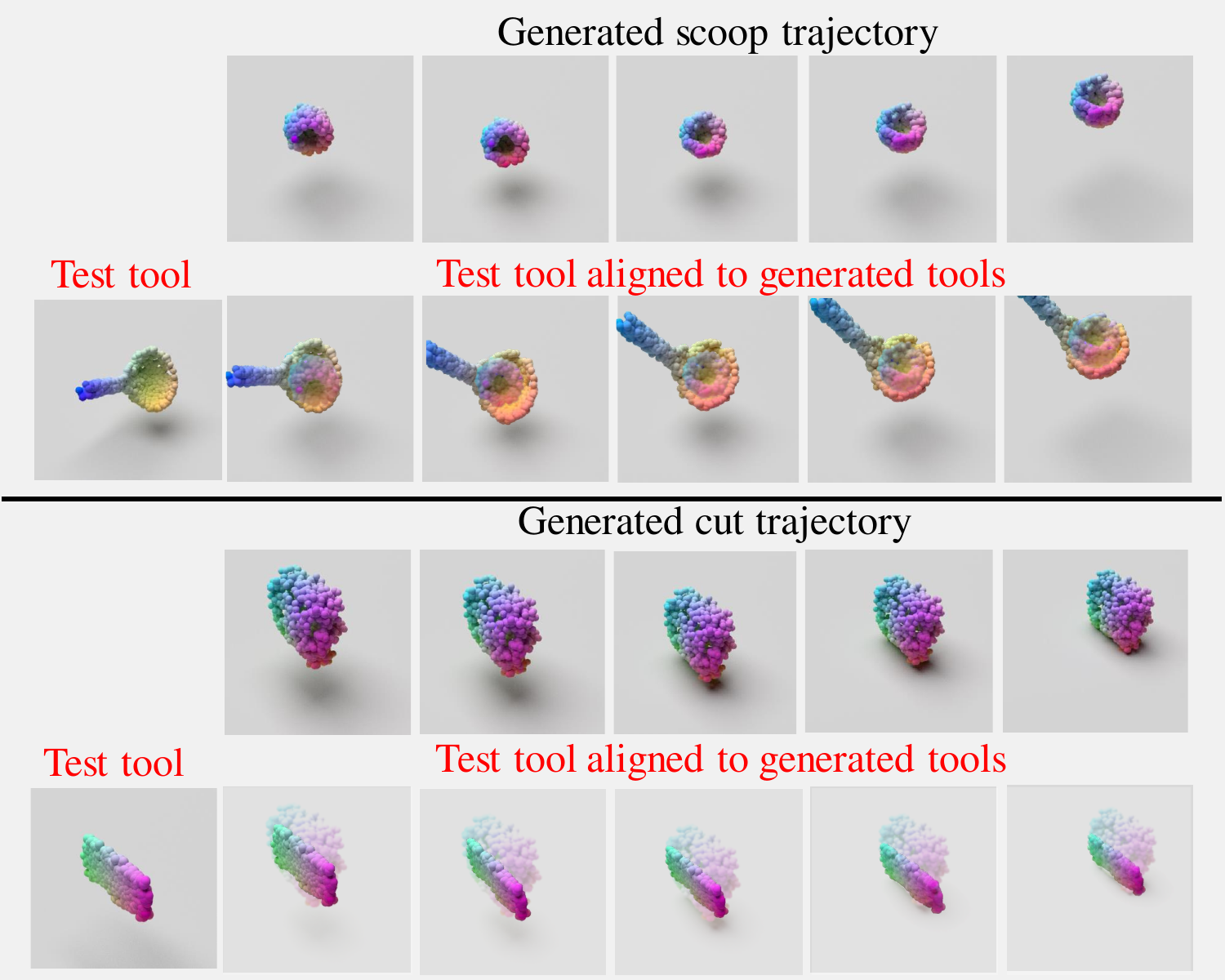}
  \caption{Examples of generated tool trajectories for scoop (top) and cut (bottom), as well as the trajectories of the test tools aligned to these generated trajectories.}
  \label{fig:genresults}
\end{subfigure}
\caption{\fig{fig:results}: Performance of all the methods across 3 settings. \fig{fig:genresults}: Examples of generated tool trajectories and test tool alignments.}
\vspace{-5mm}
\end{figure*}
\section{Results}\label{sec:results}
As shown below, we demonstrate that ToolGen is able to perform well on a variety of manipulation tasks with novel tools using just \textit{a single} model trained across multiple tasks and tools. Notably, we train with demonstrations from only one training tool per task  and we test on several unseen tools, demonstrating our method's generalization abilities. We additionally evaluate ToolGen on real world observations and use a Franka Panda robot to execute the predicted trajectory. For real world experiments, we include both the qualitative results and quantitative results to highlight our policy's effectiveness when transferred to the real world.



\subsection{Tasks and baselines}
\textbf{Tasks:} We evaluate our method against several baselines in a soft body simulator, PlasticineLab~\cite{huang2021plasticinelab}. We consider four tasks: ``Roll'', ``Cut'', ``Small scoop'' and ``Large scoop''. Example configurations and their training and test tools for these tasks are depicted in \fig{fig:tasks}. 
In our setup, all of the tools are placed far from the dough at the start of each task, as would be the case in a normal tool-use scenario. \\
\textbf{Metric:} We specify goals as 3D point clouds of different geometric shapes. We report the normalized decrease in the Chamfer Distance between the observation and the goal, computed as $s(t) = \frac{s_0 - s_H}{s_{0}},$ where $s_0, s_H$ are the initial and final Chamfer Distances to the goal respectively. To compute the performance of each method, we evaluate 10 trajectories per task per tool and then aggregate the performance across all the tasks. \\
\textbf{Baselines:} We evaluate the following baselines with different action representations. All of the baselines regress to reset transformations and delta poses, except for BC-E2E which predicts delta poses directly from the initial configuration without a reset transformation.
Details on the architectures of the baselines are described in Appendix~\ref{app:baselines} on the website.
\begin{itemize}
    \item \textbf{TrajOpt Oracle.} Differentiable trajectory optimization with ground truth dynamics from the simulator.
    \item \textbf{BC-E2E.} End-to-end behavioral cloning that outputs a $H' \times 6, (H' > H)$ vector representing the delta poses of the tool relative to the initial tool pose. Unlike the other baselines, this baseline  does not output a reset transformation.
    \item \textbf{BC-Joint.} Behavioral cloning that jointly regresses to the reset transformation and subsequent delta poses from the initial tool configuration.
    \item \textbf{BC-Latent.} Behavioral cloning that regresses to the reset transformation, moves the tool to the predicted reset pose, and then predict subsequent delta poses from a latent encoding of the scene with the tool in the reset pose. 
    \item \textbf{TFN-Traj.} Behavioral cloning that regresses to the reset transformation, moves the tool to the predicted reset pose, and then uses the updated scene to predict subsequent delta poses with the ToolFlowNet-based~\cite{seita2023toolflownet} trajectory model described in Appendix~\ref{app:pathgen} on the website. 
\end{itemize}

We examine three settings, each presenting a greater level of difficulty, detailed in \sect{sec:traintools}, \sect{sec:randinit}, and \sect{sec:unseentools}, respectively. We demonstrate that ToolGen is robust to these generalization challenges and maintains superior performance over the baselines. We additionally conduct ablation studies by removing the path generator of ToolGen, detailed in Appendix~\ref{app:ablations} on the website.

\subsection{Leveraging training tools at test time}
\label{sec:traintools}

We first test the methods on a set of held out configurations using training tools. To successfully perform the manipulation, the methods need to output the appropriate poses for the training tools 
to complete the tasks. 
\fig{fig:results} shows the performance of all the methods. We see that most methods achieve 
reasonable performance. This shows that all these methods generalize reasonably well to different goal configurations given the same training tools. In contrast, BC-E2E achieves suboptimal performance on even this simple version of the task, showing the limitations of methods that do not predict a reset transformation.
\begin{figure}[ht]
    \centering
    \includegraphics[width=0.5\textwidth]{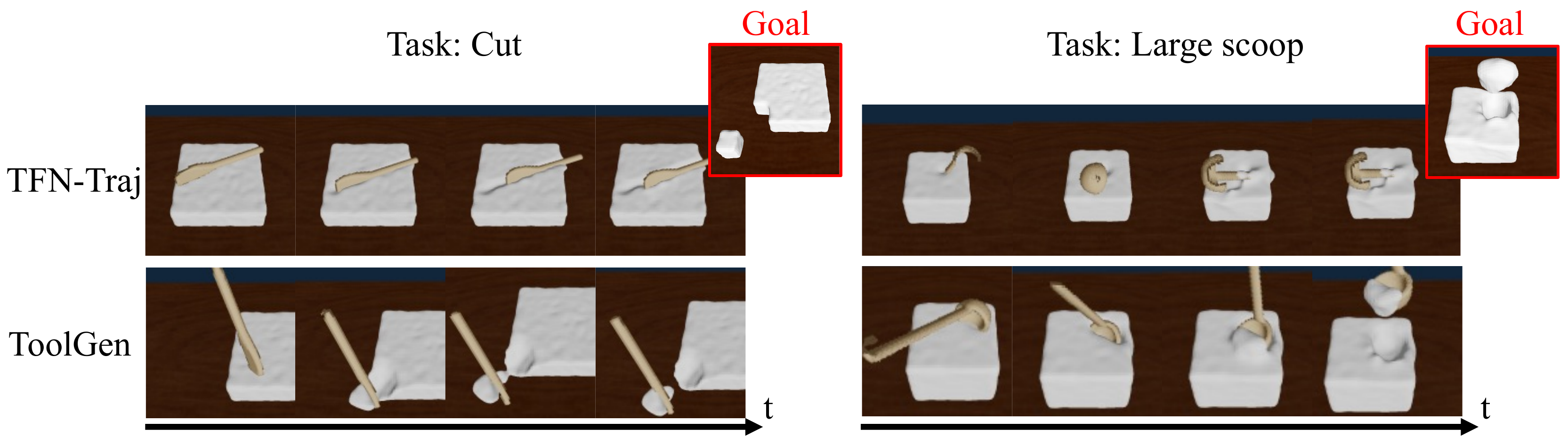}
    \caption{Example rollouts of ToolGen (ours) compared to the baseline TFN-Traj. The goal configuration of each task is shown on the top right. ToolGen can effectively use the new tool while the baseline struggles.}
    \label{fig:qualitative}
\end{figure}
\begin{figure*}[t]
    \centering
    \includegraphics[width=\linewidth]{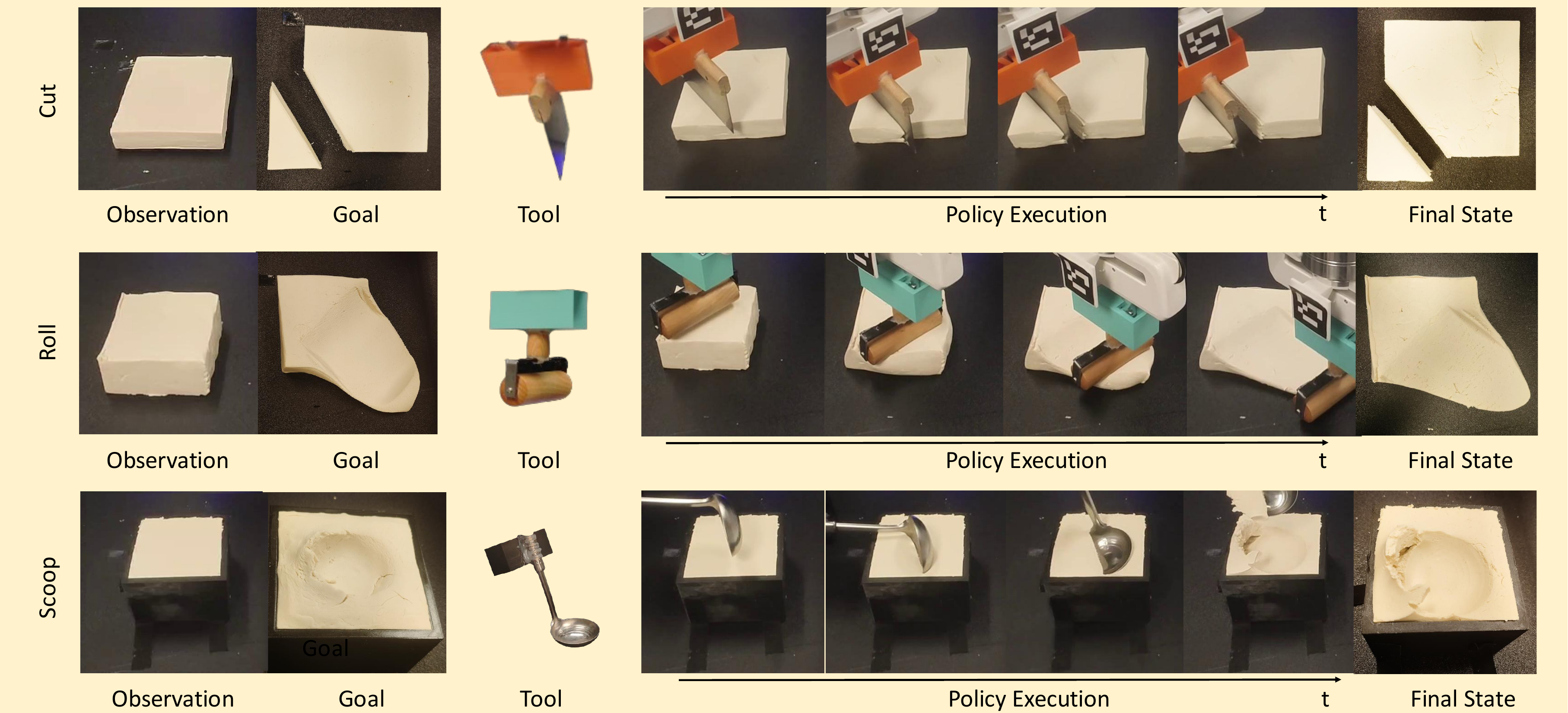}
    \caption{Results of ToolGen on real world observations for Cut (top) and Roll (middle) and Scoop(bottom). For each task, we visualize initial dough observation, the goal, the real world tool, the policy rollout of our trained policy and the final state of the policy. As a result, ToolGen can effectively generate manipulation trajectories from real world observations even though the model is trained entirely in simulation.}
    \label{fig:realworld_inference}
\end{figure*}

\subsection{Generalization to unseen initial tool poses}
\label{sec:randinit}
To simulate the fact that a tool might be in any initial configuration in the real world, we randomize the initial poses of the training tools in $SE(3)$ and rerun evaluations.
From \fig{fig:results}, we observe that ToolGen is the only method that is robust to this perturbation. Despite the fact that the baselines are trained with the same tools, they fail to generalize to unseen initial poses of the tool. On the other hand, ToolGen is robust to the initial configuration of the tool and receives no performance loss.
\subsection{Generalization to unseen tools}
\label{sec:unseentools}
\begin{table}[t]
\vspace{2mm}
\centering
\scalebox{0.7}{
\begin{tabular}{ccccc|c}
\toprule
Method & Roll & Cut & Small scoop & Large Scoop 
 & Average \\
\midrule
BC-E2E         & $0.49 \pm 0.30$     & $-0.22 \pm 0.44$  & $-0.27 \pm 0.20$ & $0.07 \pm 0.08$ & $0.02 \pm 0.26$   \\
BC-Joint       & $0.64 \pm 0.26$   & $0.00 \pm 0.09$  & $-0.05 \pm 0.10$ & $-0.01 \pm 0.07$  & $0.30 \pm 0.41$  \\
BC-Latent      & $0.70 \pm 0.15$   & $0.37 \pm 0.10$  & $-0.15 \pm 0.30$ & $0.34 \pm 0.41$ & $0.15 \pm 0.33$   \\
TFN-Traj      & $0.70 \pm 0.19$   & $0.29 \pm 0.19$  & $0.40 \pm 0.44$ & $0.35 \pm 0.40$ & $0.43 \pm 0.36$   \\
ToolGen (Ours) & $\mathbf{0.75 \pm 0.15}$   & $\mathbf{0.82 \pm 0.08}$  & $\mathbf{0.50 \pm 0.40}$ & $\mathbf{0.80 \pm 0.19}$  & $\mathbf{0.72 \pm 0.27}$  \\
\hline
\end{tabular}
}
\vspace{2mm}
\caption{Quantitative performance for different methods when using novel tools. Each value in the table represents the normalized decrease of Chamfer Distance for a specific task, measured across the use of 3 novel tools in 10 different goal configurations. The final column denotes the average performance of each method across all tasks.}
\label{tab:novel_tool}
\vspace{-3mm}
\end{table}

Finally, we evaluate the methods on a far more challenging scenario, in which our agents are given unseen tools.
We evaluate each novel tool on 10 held out goals for each task and average their performances. See \fig{fig:tasks} for a visualization of the novel tools we consider. Since the novel tools are also in arbitrary initial poses, this scenario requires the method to be robust to tool shapes as well as initial poses of the tool. \fig{fig:results} and \tbl{tab:novel_tool} shows the quantitative results of all the methods, and \fig{fig:qualitative} show examples of rollouts by ToolGen (ours) and the baseline TFN-Traj. All of the baselines fail to obtain a high performance, especially in the more challenging task of scooping (see \tbl{tab:novel_tool}). In contrast, ToolGen can leverage completely unseen tools in meaningful ways. This is because ToolGen leverages trajectory generation to alleviate the issues of distribution shift. It further uses a non-learned optimization procedure (gradient descent with multiple random initializations), which also does not suffer from a distribution shift. For more analysis, please see our Appendix~\ref{app:ablations} on the website.

We show examples of the tools generated by ToolGen (top row) as well as the test tools aligned to these generated tools (bottom row) in \fig{fig:genresults}. Overall, ToolGen achieves superior performance over the baselines in this challenging scenario of using novel tools. Remarkably, we train just a single ToolGen model across all tasks and tools, using merely one training tool per task. Despite this, ToolGen demonstrates the capacity to solve all tasks effectively when presented with novel tools.

\subsection{Inference on real world observations}
\begin{figure}[ht]
    \vspace{-0.5cm}
    \centering
    \includegraphics[width=0.5\textwidth]{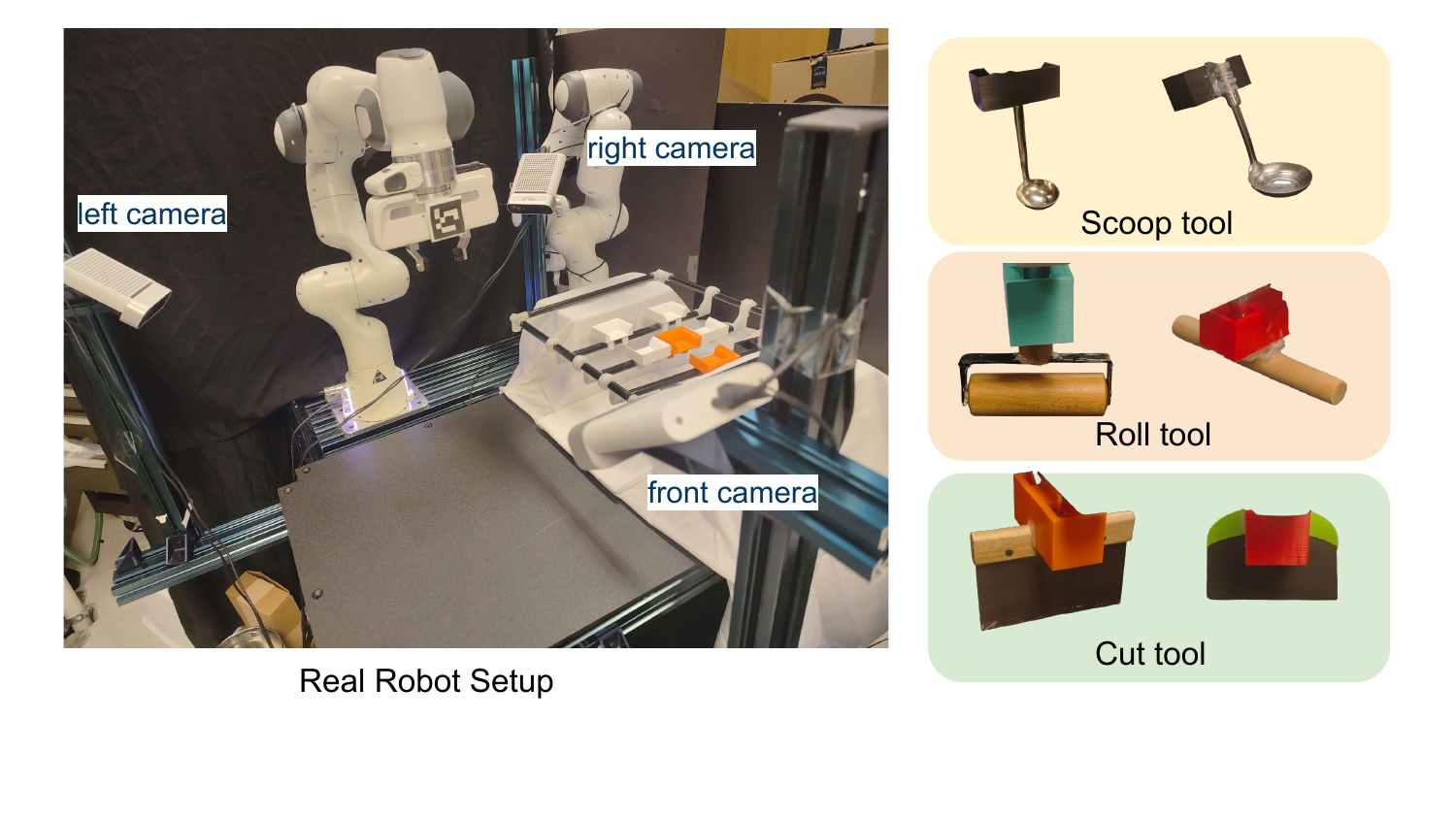}
     \vspace{-0.4in}
    \caption{On the left, it is the real world experiment setup with three cameras, one Franka Panda robot and a tool hanger. On the right, they are six tools we use for three different tasks.}
    \label{fig:realworld_setup}
   \vspace{-0.2in}
\end{figure}
For our real world experiments, we select three representative tasks, \emph{Cut}, \emph{Roll} and \emph{Scoop(Large)} to test our trained policy. In each task, we select two real world tools and attach each tool to a mount so that it can stay on the tool hanger for the robot to pick up. Details of our environment setup can be seen in Figure~\ref{fig:realworld_setup} and Appendix~\ref{sec:real_world} on the website.  
Our ToolGen model is trained entirely with simulation data. To demonstrate the robustness of ToolGen, we record point clouds of tool and dough from the real world and use ToolGen to predict the trajectory of the real world tool. To obtain the point clouds from the real world, we use three Azure Kinect cameras to record the initial dough and the tool point clouds. We then manually manipulate the dough to a desired shape and record the final point cloud as the goal point cloud. We record the point cloud of the real dough at its initial and goal states and concatenate them with the tool point cloud. The initial pose of the tool is entailed in the point cloud input. The model then output a trajectory of horizon $H=50$. In the execution stage, we use a mold to restore the dough to the recorded initial state. The robot picks up the tool and move to the recorded initial pose and executes the trajectory, where the tool first transforms from the initial pose to reset pose, and then moves through to execute the produced trajectory on the real dough. \fig{fig:realworld_inference} includes qualitative results of the robot executing the policy's rollouts. For each of the three different tasks, we show an example of the robot using one of the test tools to reshape the dough. Quantitatively, despite the gap in point cloud observations between sim and real, our method can effectively generalize to unseen tool in the real world with an average normalized decrease of Chamfer Distance of 0.77 as shown in Table~\ref{tab:real_world}. This metric shows how close the final state gets to the goal compared to the initial state of the dough. The larger normalized decrease of Chamfer Distance indicates better performance. Compared to the baseline method \emph{BC-Latent}, our method \emph{ToolGen} outperforms it on all three tasks by a large margin. \emph{BC-Latent} fails to generalize to some unseen real world tools and generates transformations that oscillate without further movements for those tools. In contrast, our method successfully generalizes to all tasks with different goals and tools. Hence, our method demonstrates smaller performance variance. To prove that our policy is comparable to human performance, two volunteers are asked to perform the same manipulation tasks as that for robots. From the table we can see that our policy's task performance is very close to humans with the largest difference of 0.08 in normalized decrease of Chamfer Distance. We also notice that \emph{Scoop} generally has worst performance compared to other two tasks because the real world dough we are using is so sticky that both the human and our trained policy struggle with detaching the scooped piece from the whole dough.
\begin{table}[ht]
\vspace{-0.2in}
\centering
\scalebox{0.7}{
\begin{tabular}{c|c|c|c|c}

\hline
 Method & Roll $\uparrow$ & Cut $\uparrow$ & Scoop $\uparrow$ & Average $\uparrow$\\
\hline
BC-Latent & $ 0.73 \pm 0.21 $ & $0.60\pm 0.46$ & $ 0.55\pm 0.12 $ & $0.57\pm 0.27$  \\
ToolGen (Ours) & $\mathbf{0.83\pm 0.09}$ & $\mathbf{0.86 \pm 0.16}$ & $\mathbf{0.63 \pm 0.14}$ & $\mathbf{0.77 \pm 0.16}$\\
\hline
Human (Oracle) & $0.91\pm 0.03$ & $ 0.90\pm 0.11$ & $0.69 \pm 0.14$ & $0.83 \pm 0.14$ \\
\hline
\end{tabular}
}
\vspace{2mm}
\caption{Quantitative results for different methods when using real world novel tools. \emph{Human Oracle} is not an automated method and serves as an upper bound for the performance of the dough manipulation tasks. Each value in the table represents the average normalized decrease of the Chamfer Distance and the standard deviation for a specific task, measured across 2 different goal configurations. Each goal configuration is tested with two different initial pose of tools. The final column denotes the average performance of each method across all tasks. The metric is computed the same way as in Table~\ref{tab:novel_tool}}
\label{tab:real_world}
\vspace{-0.2in}
\end{table}
\section{Conclusion and limitations}

In this paper, we introduce ToolGen, a novel framework for learning generalizable tool-use skills. ToolGen uses a point cloud trajectory generation approach to represent tool use and then applies sequential pose optimization for execution. This representation circumvents the issues associated with using affordances to represent tool use, and it demonstrates superior generalization capabilities, especially when evaluating on unseen test tools, given only one tool per task for training. We applied a single ToolGen model to the manipulation of deformable objects, tackling diverse tasks, goals, and tools, and we found that ToolGen significantly outperforms the baselines and generalizes effectively to many novel tools. It is our hope that ToolGen will inspire more innovative approaches for tool use representation that enable broad ranges of generalization in the future.

\textbf{Limitations:} Our method has several limitations: First, our method's execution time is considerably longer compared to that of a trained policy, due to the time needed for generating point clouds and optimizing the current tool's poses. Quantitative results are shown in Appendix~\ref{app:execution_time} on the website. We anticipate that the use of faster techniques for sequential pose optimization, such as second-order methods, could speed up our method. Secondly, as our point cloud generator is trained on limited tools, it is sometimes unable to generate accurate point clouds for novel tools and thus the alignment process could fail. A promising direction is to train on more variations of the tool to improve the generation process and make alignment easier. Further details on these failure cases are shown in Appendix~\ref{app:failures} on the website.

\section{Acknowledgement}
This work was supported by the National Science Foundation under Grant No. IIS-2046491, and the National Institute of Standards and Technology under Grant No. 70NANB23H178. Any opinions, findings, and
conclusions or recommendations expressed in this material are
those of the author(s) and do not necessarily reflect the views
of the National Science Foundation, or the National Institute of Standards and Technology. 

\bibliographystyle{bib/IEEEtran}
\bibliography{bib/conference_macro, bib/bibliography} 
\clearpage

\newpage
\appendix
\setcounter{page}{1} 


\section{Implementation Details}
\subsection{Implementation of tool scoring module}
\label{app:toolsel}
Given a set of $K$ training tools, represented as a set of point clouds, $\{P^{traintool_i}\}_{i=1:K}$, we train a tool scoring module $D_{score}$, which takes in a tool point cloud $P^{tool}$, the initial observation $P^o$, and the goal $P^g$, and it predicts a score $s$ for the tool indicating how suitable the tool is for the task. The architecture for the tool scoring module is shown in~\fig{fig:overview} (a). 
The module first encodes the tool points to a latent feature using a PointNet++~\cite{pn2} encoder. It then encodes the concatenation of observation points and goal points to another latent feature using a separate PointNet++ encoder. These latent features are concatenated and inputted through a multi-layer perceptron (MLP) to output a score.
We train the module with binary cross-entropy loss, in which the tool used in the demonstration to achieve the goal point cloud $P^g$ is considered as a positive example, and randomly selected tools from the training set are considered as negative examples. 


\subsection{Details on the path generator}
\label{app:pathgen}
The path generator $G_{path}$ starts by encoding the concatenated point clouds into a latent vector using a PointNet++~\cite{pn2} encoder. This vector is then input into a ToolFlowNet~\cite{seita2023toolflownet}-based trajectory model. The trajectory model is set to a flow dimension of $(H-1) \times 3$. The resulting output is interpreted as the tool's flow at each time step, thereby producing $H-1$ delta poses $T^{gen}_{1:H}$ via singular value decomposition~\cite{sorkine2017least, levinson2020analysis}. Finally, by utilizing this path generator with the generated tool in the reset pose $P^{gen}_0$, we create a point cloud trajectory $P^{gen}_{1:H}$. We train the path generator using the delta poses of the training tools $T_{1:H}$ as labels (from the demonstration dataset). At each timestep, we apply the ToolFlowNet~\cite{seita2023toolflownet} loss between the trajectory produced by $G_{path}$ and the actual trajectory of the training tool.
\section{Experiment Details}

\subsection{Details on tasks and demonstration data}
\label{app:dataset}
\begin{table}[ht]
    \centering
    \begin{tabular}{lcc}
    \toprule
      &  Per task & Overall  \\
    \midrule
    \# of initial configurations & 200 & 800 \\
    \# of target configurations & 200 & 800 \\
    \# of training trajectories & 180 & 720 \\
    \# of testing trajectories  & 20 & 80 \\
    \# of total trajectories & 200 & 800 \\
    \# of total transitions & $10^4$ & $4\times 10^4$ \\
    \bottomrule
    \end{tabular}
    \vspace{2mm}
    \caption{Summary of training/testing data}
    \label{tab:train-data}
\end{table}

We inherit the data generation procedure from DiffSkill~\cite{lin2021diffskill}: first, we randomly generate initial and target configurations. The variations in these configurations include the location, shape, and size of the dough and the reset pose of the tool. We then sample a specific initial configuration and a target configuration and perform gradient-based trajectory optimization to obtain demonstration data. For each task, the demonstration data consists of all the transitions from executing the actions outputted by the trajectory optimizer, and we use a task horizon of $H=50$. For each task, we perform a train/test split on the dataset and select 10 configurations in the test split for evaluating the performance for all the methods. More information about training and testing data can be found in Table~\ref{tab:train-data}.

\subsection{Details on baselines}
\label{app:baselines}
We provide additional details on each baseline below:
\begin{itemize}[leftmargin=*]
    \item \textbf{BC-E2E.} End-to-end behavioral cloning that outputs a $H' \times 6, (H' > H)$ vector representing the delta poses of the tool relative to the initial tool pose. Unlike the other baselines, this baseline  does not output a reset transformation. Here, we set $H'=60$ and use delta poses in the entire trajectory (i.e. the delta poses from interpolating the initial pose and the reset pose, as well as the subsequent delta poses during manipulation) as the label to regress on. As for the architecture, it first encodes the tool points to a latent feature using a PointNet++~\cite{pn2} encoder. It then encodes the concatenation of observation points and goal points to another latent feature using a separate PointNet++ encoder. These latent features are concatenated and inputted through an MLP to produce the delta poses (represented as a $H' \times 6$ vector).
    \item \textbf{BC-Joint.} Behavioral cloning that jointly regresses to the reset transformation and subsequent delta poses from the initial tool configuration. As for the architecture, it first encodes the tool points to a latent feature using a PointNet++~\cite{pn2} encoder. It then encodes the concatenation of observation points and goal points to another latent feature using a separate PointNet++ encoder. These latent features are concatenated and inputted through an MLP to produce the reset transformation as well as delta poses.
    \item \textbf{BC-Latent.} Behavioral cloning that regresses to the reset transformation, moves the tool to the predicted reset pose, and then predict subsequent delta poses from a latent encoding of the restrictions scene with the tool in the reset pose. As for the architecture, it first encodes the tool points to a latent feature using a PointNet++~\cite{pn2} encoder. It then encodes the concatenation of observation points and goal points to another latent feature using a separate PointNet++ encoder. These latent features are concatenated and inputted through an MLP to produce the reset transformation. For the delta poses, we encode the concatenated point clouds of the scene (observation, goal, and tool in the reset pose) into a latent vector using a PointNet++ encoder and then pass the latent feature though an MLP to produce the delta poses (represented as a $(H-1) \times 6$ vector).
    \item \textbf{TFN-Traj.} Behavioral cloning that regresses to the reset transformation, moves the tool to the predicted reset pose, and then uses the updated scene to predict subsequent delta poses for the tool with the ToolFlowNet-based~\cite{seita2023toolflownet} trajectory model described in Appendix~\ref{app:pathgen} on the webiste. 
\end{itemize}

\section{Additional Experiments}
\subsection{Ablation studies}
\label{app:ablations}

\begin{table}
    \centering
    \scalebox{0.7}{
    \begin{tabular}{cccc}
    \toprule
     Ablation Method & Training tools & Random initial pose & Novel tools \\ 
    \midrule
    ToolGen Reset w/ BC-Latent & $\mathbf{0.94 \pm 0.05}$  & $\mathbf{0.93 \pm 0.05}$ & $0.30 \pm 0.55$   \\
    ToolGen Reset w/ TFN-Traj & $0.86 \pm 0.14$  & $0.85 \pm 0.07$ & $0.36 \pm 0.60$   \\
    ToolGen (Ours) & $\mathbf{0.91 \pm 0.05}$  & $\mathbf{0.90 \pm 0.06}$ & $\mathbf{0.72 \pm 0.27}$   \\
    \bottomrule
    \end{tabular}}
\caption{Ablation results across 3 scenarios. Each value in the table
represents the normalized performance across all tasks.}
\label{tab:ablation}
\end{table}
\begin{table}
    \centering
    \scalebox{0.7}{
    \begin{tabular}{ccc}
    \toprule
      Method & Average Inference Time & Average Execution Time  \\ 
    \midrule
    TFN-Traj & $0.2$s  & $23.0$s  \\
    ToolGen (Ours) & $22.7$s  & $19.1$s \\
    \bottomrule
    \end{tabular}}
\caption{Execution times averaged for all simulation tasks.}
\label{time}
\end{table}

We conduct an ablation study on ToolGen by modifying its point cloud generator: we only generate the initial point cloud using ToolGen and align the current tool with this point cloud to determine the current tool's reset pose. Following this, we input the current tool at its reset pose into the delta pose predictors of BC-Latent and TFN-Traj to obtain the subsequent delta poses. This ablation provides a clear comparison between the process of directly regressing to the delta poses and the approach of using ToolGen to output delta poses. The performance gap between these two methods when using novel tools is displayed in Table \ref{tab:ablation}, which underscores the significance of generalization occurring in trajectory prediction. Specifically, since these two ablations regress onto the delta poses of the training tools, they tend to overfit to the training tools, causing them to produce inaccurate trajectories when faced with out-of-distribution test tools. In contrast, ToolGen inputs the generated tool into the trajectory predictors during the generation process. The generated tool minimizes the distribution shift for the path generator and thus significantly enhances the accuracy of the resulting trajectory predictions. 
\subsection{Execution time}
\label{app:execution_time}
We further compare the average inference time of ToolGen and a baseline in Table~\ref{time}. Due to the sequential pose optimization step, ToolGen requires significantly more time during inference compared to its baseline, TFN-Traj, which only requires a single forward pass in the networks. We leave improving the time efficiency of ToolGen's trajectory generation for future work.

\subsection{Failure cases}
\label{app:failures}

\begin{figure*}[ht]
    \centering
    \includegraphics[width=\linewidth]{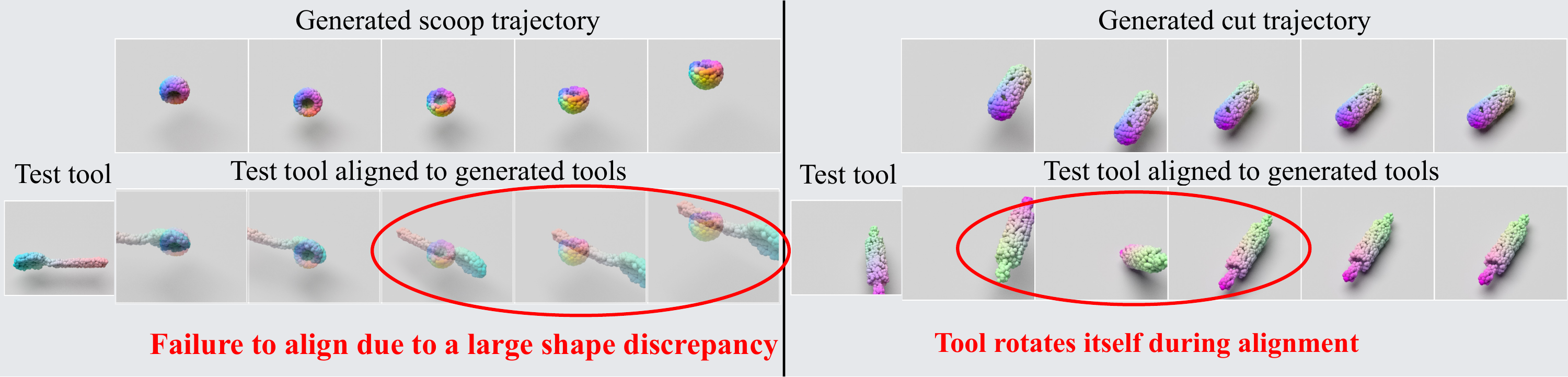}
    \caption{Example failure cases of ToolGen (ours) when trying to align the test (actual) tool with the generated tool. Left: the alignment fails due to large difference in shapes of the generated tool and the test tool. Right: occasionally during the alignment process the aligned tool would have unexpected motions.}
    \label{fig:failurecases}
\end{figure*}

In Figure \ref{fig:failurecases}, we present two typical failure scenarios that occur when trying to align a novel tool with the generated tool. The first scenario, displayed on the left of Figure \ref{fig:failurecases}, occurs when there is a substantial disparity between the generated tool (top row) and the test tool (bottom row). In this case, the optimization process fails to meaningfully align the test tool with the generated shape in the later timesteps of the trajectory. However, this issue can be alleviated by training on more diverse tool shapes, which will create a richer shape distribution for the point cloud generator to generate. \\
The second type of failure results from the optimization of delta poses. The hyper-parameter $\lambda_r$ regulates the balance between the actual alignment and the regularization of the rotation amount in delta poses, and can be sensitive to the task at hand. During our experiments, we found that a single $\lambda_r$ value generally performs well across all tasks. However, in the "Roll" task, minor problems occurred - occasionally the tool would rotate itself when aligning with the generated tool (as shown on the right of Figure \ref{fig:failurecases}). This issue can be remedied by fine-tuning the optimization's objective function and hyper-parameters for each task. For instance, by increasing the regularization parameter $\lambda_r$, we can prevent large rotations during the alignment of delta poses.
\subsection{Effects of hyperparameters}
\begin{table}
    \centering
    \scalebox{0.7}{
    \begin{tabular}{cccc}
    \toprule
     \diagbox{$\lambda_c$}{$\lambda_r$} & 0.01 & 0.1 & 0.5 \\ 
    \midrule
    0.01 & $0.45 \pm 0.30$  & $0.65 \pm 0.23$ & $0.60 \pm 0.15$   \\
    0.1 & $0.50 \pm 0.33$  & $\mathbf{0.72 \pm 0.27}$ & $0.68 \pm 0.10$   \\
    0.5 & $0.48 \pm 0.36$  & $0.49 \pm 0.29$ & $0.53 \pm 0.18$  \\
    \bottomrule
    \end{tabular}}
\caption{The effects of hyperparameters in sequential pose estimation. Each entry shows the performance cross all tasks with a particular combination of hyperparameters.}
\label{tab:hyperparam}
\end{table}
To investigate how sensitive ToolGen is to hyperparameters during sequential pose estimation, we vary $\lambda_c$ and $\lambda_r$ when optimizing for the transformations. To eliminate any stochasticity form the generation process, we only generate the     trajectories once and use the same set of generated trajectories for optimization. \tbl{tab:hyperparam} shows the performances of executing the trajectories that are optimized with different hyperparameters. In summary, $\lambda_c$, which penalizes collisions between the tool in reset pose and the environment, seems to have a greater effect than $\lambda_r$. This is because the alignment at the rest pose is being used to optimize the subsequent poses, and the error in optimizing the reset pose might cascade to the later process. We also observe that choosing a larger $\lambda_r$ will decrease the variance in performance. This is because a larger $\lambda_r$ will penalize large motions and encourage smaller and safer motions. It is worth noting that even with a large variation of these hyperparameters, ToolGen still almost always outperforms the baselines.

\subsection{Real World Experiment Details}
\label{sec:real_world}
    \subsubsection{Environment setup} 

    \quad \newline
    \indent \textbf{Robot and workspace.} The robot used for real-world execution is a 7-axis Franka Panda robot arm with a two-finger Franka Hand. The robot is fixed on a table with a $0.55m\times0.55m$ space for execution. The tool rack is placed on the left side of the execution space.
    
    \textbf{Dough.} In real world experiments, we use \textit{modeling dough}\footnote{Hygloss Products 48308 Dazzlin' Dough 3lb. White, bought from Amazon: https://www.amazon.com/Hygloss-Products-48308-Dazzlin-Dough/dp/B07SNX6BPK}
    for our manipulation tasks. For each of the three tasks, we create a mold of the same size as the dough in Taichi simulation. The real-world experiment goal states are created by human volunteers using real tools. The dough is reshaped and placed at a fixed center point before every experiment run. 
    
    \textbf{Multi-camera setup.} Here we set up multiple cameras to record dough state point clouds. Three Azure Kinect cameras are arranged around the workspace with equal distances from each other, i.e., placed in an equilateral triangle configuration, in front of the robot and on both sides of the robot, all pointing towards the geometric center of the workspace. The cameras are calibrated to form point clouds with re-projection errors less than $0.01m$. To synthesize a comprehensive view of the object, point clouds are further aligned using an Iterative Closest Point algorithm.

     \textbf{Point cloud processing.} The collected tool and dough point clouds are hollow. We interpolate them by identifying cross-sections along the x, y, or z axis and filling them with points. Then we downsample the interpolated point clouds using the same voxel size of 0.002m. This produces a uniform distribution of points, and thus allows more accurate metric calculations for the dough's target and goal point clouds.
    
    \subsubsection{Robot execution details} We use \textit{Frankapy}\footnote{https://github.com/iamlab-cmu/frankapy} as the robot controller. The delta transformations from model output is under the tool frame. To execute the trajectory with robot arm, we calculate and apply the transformation from the recorded tool frame to robot end-effector. Each target pose $(x, y, z, r, p, y)$ under robot frame is passed to Frankapy as a goal pose. Frankapy then calculates the inverse kinematics for the given goal state and execute each goal within 0.5 seconds.
    

\end{document}